\documentclass[nohyperref]{article}

\usepackage{microtype}
\usepackage{graphicx}
\usepackage{subfigure}
\usepackage{booktabs} % for professional tables
\usepackage{kotex}
\usepackage{multirow}
\usepackage{makecell}
\usepackage{hyperref}

\usepackage[accepted]{icml2022}

\usepackage{amsmath}
\usepackage{amssymb}
\usepackage{mathtools}
\usepackage{amsthm}

\usepackage[capitalize,noabbrev]{cleveref}

\theoremstyle{plain}

\theoremstyle{definition}

\theoremstyle{remark}

\usepackage[textsize=tiny]{todonotes}

\icmltitlerunning{Submission and Formatting Instructions for ICML 2022}

\begin{document}

\twocolumn[
\icmltitle{A Self-Supervised Automatic Post-Editing Data Generation Tool}




\begin{icmlauthorlist}
\icmlauthor{Hyeonseok Moon}{yyy}
\icmlauthor{Chanjun Park}{yyy,yyyc}
\icmlauthor{Sugyeong Eo}{yyy}
\icmlauthor{Jaehyung Seo}{yyy}
\icmlauthor{SeungJun Lee}{yyy}
\icmlauthor{Heuiseok Lim}{yyy}
\end{icmlauthorlist}
  
\icmlaffiliation{yyy}{Department of Computer Science and Engineering, Korea University, Seoul 02841, Korea}
\icmlaffiliation{yyyc}{Upstage, Gyeonggi-do, Korea}

% \icmlaffiliation{equal}{Corresponding author}

\icmlcorrespondingauthor{Heuiseok Lim}{limhseok@korea.ac.kr}

\icmlkeywords{Machine Learning, ICML}

\vskip 0.3in
]



\printAffiliationsAndNotice % otherwise use the standard text.

\begin{abstract}
Data building for automatic post-editing (APE) requires extensive and expert-level human effort, as it contains an elaborate process that involves identifying errors in sentences and providing suitable revisions. Hence, we develop a self-supervised data generation tool, deployable as a web application, that minimizes human supervision and constructs personalized APE data from a parallel corpus for several language pairs with English as the target language. Data-centric APE research can be conducted using this tool, involving many language pairs that have not been studied thus far owing to the lack of suitable data.
\end{abstract}

\section{Introduction}
Automatic post-editing (APE) has actively been studied by researchers because it can reduce the effort required for editing machine-translated content and contribute to domain-specific translation \cite{isabelle2007domain, chatterjee2019findings, moon2021recent}. However, APE encounters a chronic problem concerning data generation \cite{negri2018escape, lee2021adaptation}. Generally, data for the APE task comprises the source sentence (SRC), machine translation of the sentence(MT), and corresponding post-edit sentence (PE), collectively known as an APE triplet. Generating these data require an elaborate process that involves identifying errors in the sentence and providing suitable revisions. This incurs the absence of appropriate training data for most language pairs and limits the acquisition of large datasets for this purpose \cite{chatterjee-EtAl2020WMT, moon2021empirical}.

To alleviate this problem, we develop and release a noise-based automatic data generation tool that can construct APE-triplet data from a parallel corpus, for all language pairs with English as the target language. The data generation tool proposed in this study enables the application of several noising schemes, such as semantic and morphemic level noise, as well as adjustments to the noise ratio that determines the quality of the MT sentence. Using this tool, the end-user can generate high-quality APE triplets as per the intended objective and conduct data-centric APE research.

\section{Data Construction Process and Tool Implementation}
\begin{figure*}[t]
  \centering
  \includegraphics[scale=0.45]{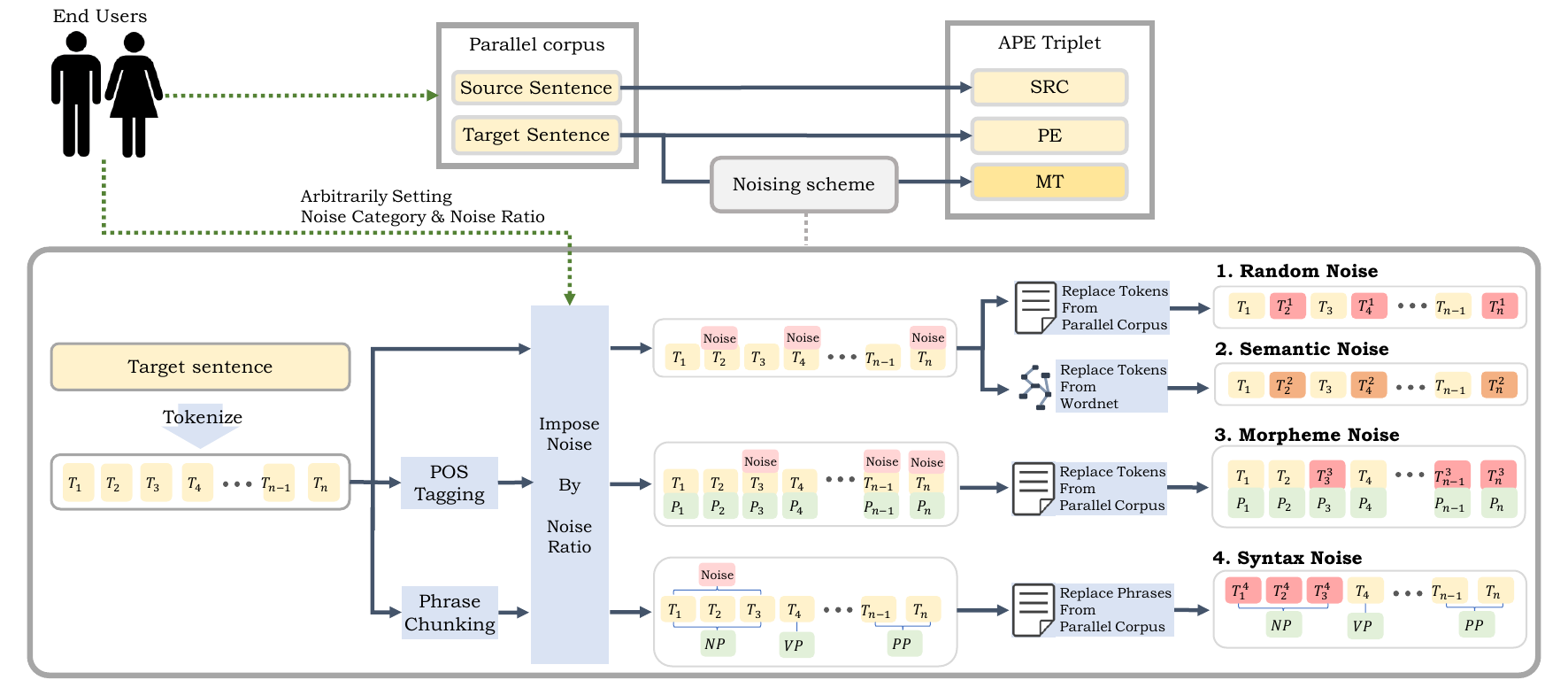}
  \caption{Overview of data construction process. $T_i$ refers to the tokenized component of the target sentence, $P_i$ indicates the POS tag corresponding to token $T_i$, and $T_{i}^{j}$ refers to the replacement token generated from the $j^{th}$ noise category. Throughout this process, the end-user can arbitrarily set the noise category and noise ratio and thereby obtain personalized APE triplets.}
  \label{fig:process}
\end{figure*}

\paragraph{Process} 
We developed an APE data generation tool that automatically construct APE datasets from a given parallel corpus. The working of our tool is outlined in Figure \ref{fig:process} and described as follows. The source and target sentence in the parallel corpus are considered the SRC and MT of the APE triplet, respectively, and a noising scheme is implemented for the generation of a pseudo-MT \cite{lee2020noising}. Noise is introduced by replacing certain tokens in the target sentence with others, using one of the four following noising schemes. 

\begin{itemize}

\begin{table*}[h]
\caption{Performance of the APE models trained with augmented training data, generated by following our approaches.}
\label{tab:performance}
\vskip 0.15in
\centering
\begin{small}
\begin{sc}
\begin{tabular}{ccccccc}
\toprule
\multirow{2}{*}{Test Set} & \multicolumn{2}{c}{Amazon} & \multicolumn{2}{c}{Microsoft} & \multicolumn{2}{c}{Google} \\
{} & SacreBLEU & TER & SacreBLEU & TER & SacreBLEU & TER \\
\midrule
Baseline & 22.192 & 59.787 & 25.130 & 59.287 & 33.115 & 51.928 \\
RANDOM & 42.317 & 44.919 & 42.287 & 44.853 & 42.468 & 44.724 \\
SEMANTIC & 42.121 & 45.208 & 42.120 & 45.157 & 42.358 & 44.915 \\
MORPHEMIC & 42.857 & 44.097 & 42.893 & 44.130 & 42.966 & 44.078 \\
SYNTACTIC & 42.725 & 44.346 & 42.732 & 44.326 & 42.847 & 44.212 \\
\bottomrule
\end{tabular}
\end{sc}
\end{small}
\vskip -0.1in
\end{table*}

\item {\textsc{Random}: ~~} The random noising scheme replaces tokens in the original target sentence in a random manner \cite{park2020neural}. In this scheme, no semantic or syntactic information is reflected, and the noise is applied simply by replacing existing tokens with others from the target side of the parallel corpus.

\item {\textsc{Semantic}: ~~ } In the semantic noising scheme, each token in the target sentence is replaced with the corresponding synonym retrieved from the WordNet database \cite{fellbaum2010wordnet}. As all the tokens are replaced with semantically identical words, the APE model can learn to correct instances of inappropriate word-use arising from subtle differences in context or formality.

\item{\textsc{Morphemic}: ~~ } In the morphemic noising scheme, certain tokens in the sentence are replaced using tokens with the same part-of-speech (POS) tag. The replacement token is extracted from the given parallel corpus.

\item{\textsc{Syntactic}: ~~ } The syntactic noising scheme implements phrase-level substitutions. 
Prior to the noising process, phrase chunking is performed using begin, inside, outside (BIO) tagging, and MT is created via replacement with an identically tagged phrase.
\end{itemize} 

\paragraph{Tool Implementation}
For the implementation of our tool, end-users need to specify the intended noise category and noising ratio and provide a parallel corpus with corresponding language pairs. The proposed tool is distributed as a web application developed using the Flask framework \cite{grinberg2018flask}. For the implementation of the noising process, Natural Language Toolkit (NLTK) \cite{bird2009natural} and SENNA NLP toolkit\footnote{https://ronan.collobert.com/senna/license.html} are utilized. In particular, NLTK is used for POS tagging and WordNet retrieval in the morphemic and semantic noising schemes, whereas SENNA is utilized for BIO tagging in the syntactic noising scheme. The web application of the proposed tool is publicly available \footnote{\url{http://nlplab.iptime.org:9092/}}.

\section{Experimental Results}
To inspect the effectiveness of our tool, we train APE models with training corpora obtained by each data augmentation methodologies and inspect the performance of each model. For implementing these, we adopt APE SOTA approcah \cite{yang2020hw} that fine-tuning APE task to pre-trained NMT model. Detailed experimental settings are the same as proposed in \citet{moon2022automatic}. Experimental results are depicted in Table~\ref{tab:performance}.

As shown in results, we can obtain high-performance APE models only with our augmented data, without human-revised data. These results shows that our approaches relieve the needs for the high-level expert human labor required in APE data generation an d can generate high quality APE data with parallel sources. This can promote universal research to low-resource language pairs that official APE data has not been released.

\section{Conclusion}
The tool proposed in this paper reduces the need for expert-level human supervision generally required for APE data generation, thereby facilitating APE research on many language pairs that have not been studied thus far. The personalization capability of the proposed APE data generation tool can enable data-centric APE research that derives optimal performance through high-quality data.

\section*{Acknowledgment}
This research was supported by the MSIT(Ministry of Science and ICT), Korea, under the ITRC(Information Technology Research Center) support program(IITP-2018-0-01405) supervised by the IITP(Institute for Information \& Communications Technology Planning \& Evaluation) and Basic Science Research Program through the National Research Foundation of Korea(NRF) funded by the Ministry of Education(NRF-2021R1A6A1A03045425). Heuiseok Lim$^\dagger$ is a corresponding author.

\nocite{langley00}

\bibliography{example_paper}
\bibliographystyle{icml2022}

%%%%%%%%%%%%%%%%%%%%%%%%%%%%%%%%%%%%%%%%%%%%%%%%%%%%%%%%%%%%%%%%%%%%%%%%%%%%%%%
%%%%%%%%%%%%%%%%%%%%%%%%%%%%%%%%%%%%%%%%%%%%%%%%%%%%%%%%%%%%%%%%%%%%%%%%%%%%%%%
% APPENDIX
%%%%%%%%%%%%%%%%%%%%%%%%%%%%%%%%%%%%%%%%%%%%%%%%%%%%%%%%%%%%%%%%%%%%%%%%%%%%%%%
%%%%%%%%%%%%%%%%%%%%%%%%%%%%%%%%%%%%%%%%%%%%%%%%%%%%%%%%%%%%%%%%%%%%%%%%%%%%%%%
% \newpage
% \appendix
% \onecolumn

% You can have as much text here as you want. The main body must be at most $8$ pages long.
% For the final version, one more page can be added.
% If you want, you can use an appendix like this one, even using the one-column format.
%%%%%%%%%%%%%%%%%%%%%%%%%%%%%%%%%%%%%%%%%%%%%%%%%%%%%%%%%%%%%%%%%%%%%%%%%%%%%%%
%%%%%%%%%%%%%%%%%%%%%%%%%%%%%%%%%%%%%%%%%%%%%%%%%%%%%%%%%%%%%%%%%%%%%%%%%%%%%%%

\end{document}